\documentclass[a4paper,conference]{IEEEtran}

\usepackage{cite}
\usepackage{multirow}
\usepackage{amssymb}
\usepackage{doi}

\usepackage{xcolor}

\ifCLASSINFOpdf
  \usepackage[pdftex]{graphicx}
   \DeclareGraphicsExtensions{.pdf,.png}
\else
\fi

\usepackage{amsmath}
\usepackage{array}

\usepackage{url}

\begin{document}

\title{Fast Approximate Modelling of the Next Combination Result for Stopping~the~Text Recognition in a Video}

\author{\IEEEauthorblockN{Konstantin Bulatov}
\IEEEauthorblockA{Smart Engines Service LLC, \\
FRC CSC RAS, Moscow, Russia\\
Email: kbulatov@smartengines.com}
\and
\IEEEauthorblockN{Nadezhda Fedotova}
\IEEEauthorblockA{Smart Engines Service LLC, \\
Moscow, Russia\\
Email: nfedotova@smartengines.com}
\and
\IEEEauthorblockN{Vladimir V. Arlazarov}
\IEEEauthorblockA{Smart Engines Service LLC\\
MIPT, Moscow, Russia \\
Email: vva@smartengines.com}}


\maketitle

\begin{abstract}
In this paper, we consider a task of stopping the video stream recognition process of a text field, in which each frame is recognized independently and the individual results are combined together. The video stream recognition stopping problem is an under-researched topic with regards to computer vision, but its relevance for building high-performance video recognition systems is clear.
		
Firstly, we describe an existing method of optimally stopping such a process based on a modelling of the next combined result. Then, we describe approximations and assumptions which allowed us to build an optimized computation scheme and thus obtain a method with reduced computational complexity.
		
The methods were evaluated for the tasks of document text field recognition and arbitrary text recognition in a video. The experimental comparison shows that the introduced approximations do not diminish the quality of the stopping method in terms of the achieved combined result precision, while dramatically reducing the time required to make the stopping decision. The results were consistent for both text recognition tasks.
\end{abstract}


\IEEEpeerreviewmaketitle

\section{Introduction}

Video processing has become a rich and dynamic branch of the research in the computer vision field. The problems of video stream analysis include object detection and segmentation~\mbox{\cite{blazingly_fast_object_segmentation_chen, monet_xiao}}, object tracking \cite{detect_and_track_girdhar, a_prior_less_method_for_tracking_lin}, super-resolution~\cite{deep_video_super_resolution_jo}, text recognition \cite{efficient_video_scene_text_spotting_arxiv} and many more. Modern computer vision applications employ methods of automated video processing not only for high-end industrial vision systems but for mobile devices such as smartphones or tablet computers as well.
	
One of the most relevant computer vision problems is automated data entry by means of text recognition. Text detection and recognition have applications in such fields as business processes automation, road traffic monitoring, government services, mobile payments~\mbox{\cite{business_cards_reader, a_survey_on_ocr_arxiv, ravneet}}, life augmentation for people with disabilities~\cite{icmv-jabnoun} and more. The research is targeted on improving the speed and accuracy of camera-based information extraction, given particular challenging conditions, such as poor illumination, low camera quality \cite{brno}, optical distortions and noise, poor focus and motion blur \cite{ngoc_fabrizio_geraud_icdarw}. Each year new and improved methods for arbitrary scene text detection recognition are published \cite{8545198, 8545047, 9010273}, including the works which focus on processing text in videos \cite{10.1145/3343031.3351093, Cai_robust_detect}. An important requirement in many applications of text recognition is for the system to be able to operate in real time, which is especially relevant for such use cases as recognition of road scene objects such as traffic signs~\cite{ieee2015greenhalgh}, assisting the visually impaired~\mbox{\cite{10.1007/978-3-030-03801-4_44, mobile_ocr_without_sight, 5711544}}, and others.


\begin{figure}[!t]
\centering
    \includegraphics[width=0.7\linewidth]{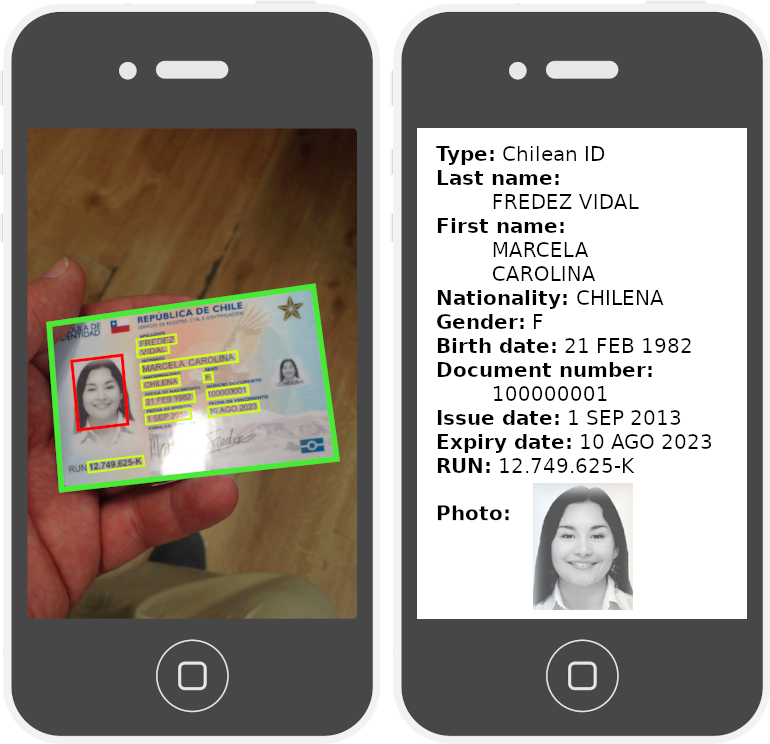}
	\caption{Illustration of identity document recognition and data extraction on a mobile device}
	\label{fig:phone_image}
\end{figure}

A special case of text recognition is represented within mobile identity document recognition systems. The recognition of identity documents is encumbered by specific features of such documents, e.g. textured background, holographic security elements which are obstructing the text recognition, reflective document surfaces which is prone to highlights, etc.~\cite{midv500}. At the same time, an important aspect of identity document recognition systems is their low error tolerance -- the cost of recognition mistakes are high, as the recognized data is then used for personal identification, government services, financial transactions and in other sensitive fields. The scope of computer vision problems related to identity documents recognition includes document detection and location~\mbox{\cite{ngoc_fabrizio_geraud_das, skoryukina_documents_location}}, document layout analysis~\cite{povolotskiy2019dynamic}, face detection~\cite{l3i_face_detection}, and, of course, text fields recognition~\cite{SMART_IDREADER_ICDAR, RYAN2015520, recstr_access}. \figurename~\ref{fig:phone_image} illustrates the use case for information extraction from an identity document using a mobile device camera.

Using video input in recognition systems presents an opportunity to reduce the text recognition errors and thus increase the information extraction reliability, both in the context of arbitrary text recognition or in a more specialized case of document fields recognition. Combination of multiple recognition results of the same object obtained from different frames has been shown to be an effective way of improving the recognition accuracy in a video \cite{vestnik_integration, SMART_IDREADER_ICDAR}. This approach, however, gives rise to another problem -- how to decide when there is enough accumulated information and the video stream recognition process should be stopped. Without access to the ground truth, the stopping methods should be able to make a decision whether the result could be improved or not and whether it is justified to spend more time to recognize additional frame in an effort to improve the combined result. The stopping rules are particularly crucial for real-time tracking and recognition of multiple objects, such as vehicle license plates or traffic signs~\cite{ieee2015greenhalgh}.

However, after extensive research, we have found that the problem of optimal stopping is left almost unexplored in the field of computer vision, despite its importance for video processing. At the same time, optimal stopping is a known problem in the field of decision theory, mathematical statistics, and economics, and new theoretical results continue to be produced for its different variations~\mbox{\cite{The-Monotone-Case-Approach, ferguson_house_hunting}}. A few methods have been proposed for the problem of stopping the video stream recognition process~\cite{stopper_slavin, Bulatov2019} in the context of document recognition. The method proposed in~\cite{stopper_slavin} is based on the clustering of the input results sequence and making the stopping decision based on the statistical characteristics of the obtained clusters. The stopping method described in~\cite{Bulatov2019} is based on the modelling of the next combined result and making the stopping decision based on the estimated expected distance between the current combined result and the next one. The latter method was tested on text recognition results using both Levenshtein-based and end-to-end text string distance metrics and exhibited higher effectiveness in comparison with the clusters analysis method~\cite{stopper_slavin}. The method was also tested for text recognition result model with per-character class membership estimations and also proved to be more effective~\cite{bulatov2019integrated} in comparison with the other methods. 
	
All the related works considered only the combined result accuracy characteristics and the achieved mean number of processed frames, without any attention to the time required to compute the necessary estimations on which the stopping decision is based. In particular, the process of modelling of the combined result at the next stage, proposed in \cite{Bulatov2019} has a high computational complexity which could diminish the positive effects of the stopping method, especially if executed in real time on a mobile device. The goal of this paper is to construct an approximate method of modelling of the next combined result required to make the stopping decision according to the method described in \cite{Bulatov2019}, which would have reduced computational complexity, and evaluate the constructed methods for the tasks of arbitrary text recognition as well as document fields recognition in videos.
	
Section \ref{sec:existing_method} provides an overview of the studied stopping method. The method was originally proposed in \cite{Bulatov2019} and further evaluated in \cite{bulatov2019integrated}. In section \ref{sec:proposed_optimizations} the proposed approximations and optimizations are described. The experimental evaluation of the base stopping method and the proposed optimizations is presented in the final section \ref{sec:experiments}.

\section{Existing method description} \label{sec:existing_method}

In this section, we will provide a detailed description of the procedure of combining the individual per-frame text string recognition results using the algorithm described in~\cite{vestnik_integration}, and stopping the video stream recognition process using the method described in~\cite{Bulatov2019, bulatov2019integrated}.

A text string recognition result $X$ with per-character alternatives can be represented as a matrix:
\begin{equation}
\label{eq:ocrstring}
X = (x_{jk}) \in [0.0, 1.0]^{M \times K}, \quad \forall j~~\sum\limits_{k=1}^K x_{jk} = 1.0,
\end{equation}
where $M$ is the length of the string (number of characters) and $K$ is the number of character classes. Each row~$(x_{j1}, x_{j2},\ldots, x_{jK})$ of the matrix represents the classification result for each individual character and contains membership estimations for each class. The value of each membership estimation is a real number from the range~$[0.0, 1.0]$, and the sum of all membership estimations for each given character classification result equals to $1.0$. The combined recognition result of the text string in a video stream is represented with the same data structure. 

\begin{figure}
	\centering
		\includegraphics[width=0.9\linewidth]{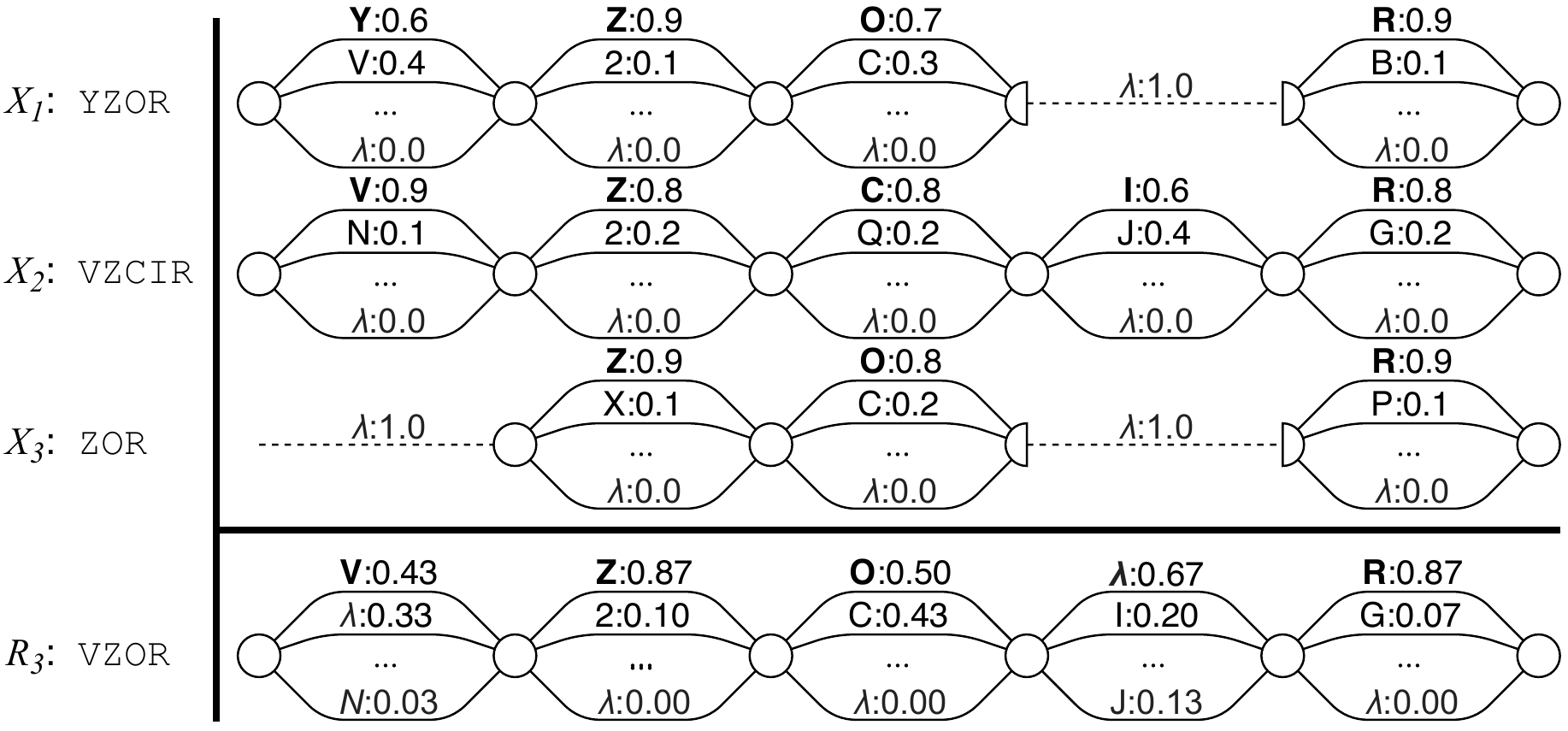}
	\caption{An illustration of per-frame text string recognition results alignment and combination}
	\label{fig:rover_illustration}
\end{figure}

During the recognition process, we observe a sequence of text string recognition results $X_1, X_2, \ldots$. The goal is to produce a combined result with the highest accuracy, that is, the closest possible to the correct text string value~$X^*$ in terms of some predefined metric. In a more general case, there is also a non-negative weight $w_i$, associated with each~$X_i$, which represents the desired contribution of the result~$X_i$ in the combination.

On stage $n$, after obtaining the observation $X_n$, it is combined with the previously accumulated recognition results as follows. Firstly, each character classification result is expanded with an ``empty'' class label $\lambda$ with membership estimation~$0.0$. In terms of the matrix representation, this corresponds to adding a zero-valued column at the beginning of the matrix~\eqref{eq:ocrstring}. Then an alignment is calculated between~$X_n$ and the previously obtained combined result $R_{n-1}$ using a dynamic programming procedure to determine the optimal matching between rows of $X_n$ and $R_{n-1}$. After the character alignment is determined the corresponding character classification results are combined by calculating a weighted average of membership estimations for each class, and using an ``empty'' classification result (with class $\lambda$ having membership estimation $1.0$) for pairing with unaligned characters. 

\figurename~\ref{fig:rover_illustration} illustrates the alignment and the combination result for three text string recognition results. For ease of visual representation, each frame result is represented as a weighted identity transducer \cite{LLOBET} with each character classification result corresponding to a set of labelled transitions between two consequent states.

After the combination is performed and the result $R_n$ is obtained, the stopping method is applied to make a decision whether $R_n$ should be returned as a final result or whether the process should continue (i.e. the observation $X_{n+1}$ should be acquired). Stopping method introduced in \cite{Bulatov2019} operates under the assumption that the expected distances between two consecutive combined results do not increase from stage to stage. Under this assumption the problem can be viewed as a monotone stopping problem, at least starting from a certain stage, and an optimal stopping rule for it should behave in a ``myopic'' way, i.e. make the decision as if the next stage of the process will be the last one. To approximate an optimal stopping rule on stage $n$ the expected distance is calculated from the current combined result $R_n$ to the next~$R_{n+1}$. The process is stopped if this expected distance is not higher than the predefined threshold which represents the cost of each observation.

The proposed method of calculation of the expected distance to the next combined result is to perform modelling of the next result by sampling already accumulated observations as candidates for the next one:
\begin{equation}
\label{eq:modelling_estimation}
\hat{\Delta}_n = \cfrac{1}{n+1}\left(\delta + \sum\limits_{i=1}^n \rho\left(R_n, R(X_1, \ldots, X_n, X_i\right)\right),
\end{equation}
where $n$ is the stage number, $\delta$ -- external parameter, $\rho$ is a metric function defined on the set of text string recognition results, $R_n$ is a combined result obtained on the $n$-th stage, and~$R(X_1, \ldots, X_n, X_i)$ is the combined result obtained by testing the individual frame recognition result $X_i$ as a candidate for the next observation.

For the validity of the stopping methods as it is described in~\cite{Bulatov2019} any metric function $\rho$ between the text string recognition results can be used, provided that it satisfies the triangle inequality. For experiments in \cite{Bulatov2019} and \cite{bulatov2019integrated} a normalized version of the Generalized Levenshtein Distance (GLD) \cite{ngld_yujian} was used. For calculating GLD a metric $\rho_C$ must be defined on the set of individual character classification results (on individual rows of the matrices). For this purpose a scaled taxicab metric can be used:
\begin{equation}
\label{eq:scaled_taxicab}
\rho_C(a, b) = \cfrac{1}{2}\sum\limits_{k=0}^K | a_k - b_k |,
\end{equation}
where $a$ and $b$ are two matrix rows, $a_0$ and $b_0$ are the respective membership estimations for the ``empty'' class $\lambda$, and $a_k$ and~$b_k$ are the respective membership estimations for character classes in the alphabet for all $k>0$.

The described stopping method was tested on text string recognition results without membership estimations \cite{Bulatov2019}, using ROVER \cite{FISCUS-659110} as a combination algorithm. Later it was tested with another per-frame recognition method and with an extended text string recognition result model \cite{bulatov2019integrated} using a combination algorithm \cite{vestnik_integration} based on a ROVER approach. In both experiments, it was shown that using such an approach for a given average number of processed frames the lower error level could be achieved and vice versa.

However, the obvious downside of this stopping method is the complexity of the decision making algorithm, in particular the time required to compute the expected distance estimation~\eqref{eq:modelling_estimation}. Given two text string recognition results $X$ and $Y$ with lengths $|X|$ and~$|Y|$ the complexity of their combination using algorithm \cite{vestnik_integration} is $O(|X|\cdot|Y|\cdot K)$, as the alignment procedure requires $O(|X|\cdot|Y|)$ calculations of the individual character classification metric function~$\rho_C$~\eqref{eq:scaled_taxicab}. For the same reason, the complexity of calculating GLD~$\rho(X, Y)$ is also $O(|X|\cdot|Y|\cdot K)$. Consider~$M$ as the maximal length of individual results $X_i$, and $S_n$ as the length of combined result~$R_n$ (both are in terms of the number of rows). The complexity of performing each test combination required for computing the sum in \eqref{eq:modelling_estimation} is~$O(S_nMK)$ and the worst-case estimation for the length of the next combined result candidate is $O(S_n+M)$. Thus, to compute the distance sample from the current combined result to the next, another~$O(S_nK(S_n+M))$ operations have to be performed. The total aggregate complexity of computing the expected distance estimation $\hat{\Delta}_n$ \eqref{eq:modelling_estimation} on stage $n$ is $O(nS_nK(S_n+M))$.

\section{Proposed optimizations} \label{sec:proposed_optimizations}
	
In this section we will introduce approximations which would help us to compute the approximate value of the estimation \eqref{eq:modelling_estimation} more efficiently.

\subsection{General approximations} \label{sec:approximations}

To obtain a more efficient method for computing the expected distance estimation \eqref{eq:modelling_estimation} let us introduce the following approximations: 

{\bf Approximation 1:} naive alignment. During the test combination $R(X_1, \ldots, X_n, X_i)$ the candidate frame recognition result $X_i$ has to be aligned with $R_n$, which already contains $X_i$ (as it was aligned with $R_{i-1}$ on the $i$-th stage of the process). As an approximation of this alignment, we will assume that rows of~$X_i$ will be aligned with the same rows of~$R_n$ with which the corresponding components were combined on stage $i$. This coarse assumption will allow to skip the costly alignment for each test combination, as the row indices of $R_n$ with which the rows of $X_i$ should be aligned are known in advance.
	
{\bf Approximation 2:} naive Levenshtein. Given approximation~1 the length of the combined result $R(X_1, \ldots, X_n, X_i)$ stays they same as the length of~$R_n$. We will then assume that alignment of $R_n$ and $R(X_1, \ldots, X_n, X_i)$, which is required to compute the GLD between them, is direct, i.e. each $j$-th row of $R_n$ is aligned with $j$-th row of $R(X_1, \ldots, X_n, X_i)$. Thus, the distance between $R_n$ and $R(X_1, \ldots, X_n, X_i)$ is a sum of distances in terms of the scaled taxicab metric~$\rho_C$ \eqref{eq:scaled_taxicab} between their rows with the same index. 

To quickly compute the approximate value of an expected distance estimation \eqref{eq:modelling_estimation}, during the combination of input results~$X_1, X_2, \ldots$ we will maintain and update a three-dimensional matrix $Y_n$:
\begin{equation}
\label{eq:matrix_Y}
Y_n = (y_{ijk}) \in [0.0, 1.0]^{n \times S_n \times (K + 1) },
\end{equation}
such that $i$ is the index of input per-frame result, $j$ is the index of a row of the current combined result, $k$ is the character class index, $y_{ijk}$ is a membership estimation of class $k$ from a row of input $X_i$ which was aligned and merged into the~$j$-th row of the current combined result $R_n$.

Given the approximations 1 and 2, using a GLD as the metric function $\rho$, for $j$-th component of the combined result $R_n$ and for each individual character class index~\mbox{$k \in \{0, 1, \ldots, K\}$} ($k=0$ being the index of an ``empty'' class label $\lambda$), considering $X_i$ as a candidate for the next frame recognition result adds the following contribution to the sum of distances in \eqref{eq:modelling_estimation}:
\begin{equation}
\label{eq:weighted_contribution}
\Delta_{ijk} = \cfrac{1}{2} \left| \cfrac{A_{jk}}{W} - \cfrac{A_{jk} + y_{ijk} w_i}{W + w_i} \right|,
\end{equation}
where $w_i$ is the weight associated with an input per-frame result $X_i$; $A_{jk}$ is the weighted sum of membership estimations of class $k$ corresponding to $j$-th component of the combined result: \mbox{$A_{jk}=\sum_{i=1}^n y_{ijk} w_i$}; and $W$ is the sum of all weights:~\mbox{$W=\sum_{i=1}^n w_i$}.
	
The approximation of the expected distance estimation~\eqref{eq:modelling_estimation} can now be computed as follows:
\begin{equation}
\label{eq:weighted_coarse_modelling}
\hat{\Delta}_n \approx \cfrac{1}{n+1} \left(\delta + \sum\limits_{i=1}^n \sum\limits_{j=1}^{S_n} \sum\limits_{k=0}^K \Delta_{ijk}\right).
\end{equation}
	
Since for all $j$ and $k$ the weighted sum $A_{jk}$ of membership estimations can be computed on-the-fly during combination of $X_n$ and $R_{n-1}$, the approximate estimation \eqref{eq:weighted_coarse_modelling} can be computed with complexity $O(nS_nK)$.

\subsection{Unweighted case} \label{sec:unweighted_case}

Often there are no weights $w_i$ associated with input text string recognition results $X_i$, i.e. all input recognition results have an equivalent contribution to the final combined result. In this case the expression \eqref{eq:weighted_contribution} can be further simplified:
\begin{equation}
\label{eq:unweighted_contribution}
\Delta_{ijk} = \cfrac{1}{2} \left| \cfrac{A_{jk}}{n} - \cfrac{A_{jk} + y_{ijk}}{n+1} \right| = 
\cfrac{\left| A_{jk} - n\cdot y_{ijk} \right|}{2n(n+1)}.
\end{equation}
	
If $A_{jk}$ are precalculated the three sums in the approximate estimation scheme \eqref{eq:weighted_coarse_modelling} are independent, thus the higher-level summation across the input sequence can be brought to the lowest level. The sum $\sum_{i=1}^n \Delta_{ijk}$ can then be computed with complexity lower than $O(n)$. Consider~$L_{jk}\subset \{1, 2, \ldots, n\}$ as a subset of indices such that~\mbox{$\forall i\in L_{jk}:~n\cdot y_{ijk} < A_{jk}$}. Let $B_{jk}$ denote the sum of elements with such indices:~\mbox{$B_{jk}=\sum_{i\in L_{jk}} y_{ijk}$}. By performing separate summations across indices in $L_{jk}$ we can remove the absolute value bars in the expression~\eqref{eq:unweighted_contribution}:
\begin{multline}
\label{eq:unweighted_contribution_summation}
\sum\limits_{i=1}^n \Delta_{ijk} = \cfrac{1}{2n(n+1)}\sum\limits_{i=1}^n \left| A_{jk} - n \cdot y_{ijk}\right | = \\ =	\cfrac{1}{2n(n+1)} \biggl(|L_{jk}|\cdot A_{jk} - n\cdot B_{jk} + \\ + n \cdot (A_{jk} - B_{jk}) - A_{jk} \cdot (n - |L_{jk}|) \biggr) = \\ =
\cfrac{1}{n(n+1)}(A_{jk}\cdot |L_{jk}| - n\cdot B_{jk}).
\end{multline}

Thus, to efficiently compute \eqref{eq:unweighted_contribution_summation} we need to be able to quickly calculate the values of $|L_{jk}|$ and $B_{jk}$ for each~$j$ and~$k$. In order to do that we need to set up a data structure for $y_{1jk}, y_{2jk}, \ldots, y_{njk}$ which supports fast insertion (which is required when $Y_{n-1}$ is updated to incorporate~$X_n$ and produce~$Y_n$) and fast computation of the quantity $|L_{jk}|$ and sum $B_{jk}$ of elements lower than the average. For this purpose we can use balanced binary search trees such as treaps \cite{Blelloch:1998:FSO:277651.277660}, with which both insertion and the queries would require $O(\log n)$ operations. The complexity of computing the approximate expected distance estimation \eqref{eq:weighted_coarse_modelling} is thus reduced to~$O(S_n K \log n)$. 
	
For large $n$ this computation scheme is significantly more efficient than the direct summation \eqref{eq:weighted_coarse_modelling}, however for small values of $n$ it could be impractical to implement due to high computational overhead relative to the number of elements in each binary search tree. Thus in the experimental section \ref{sec:experiments} we will evaluate both computational schemes.
	
\subsection{Using a normalized GLD} \label{sec:ngld}

Optimizations proposed in subsections \ref{sec:approximations} and \ref{sec:unweighted_case} consider a GLD as the distance function $\rho$. In this subsection we will consider a normalized version of the GLD \cite{ngld_yujian}, which was used for measuring a character-level recognition error rate in~\cite{Bulatov2019} and \cite{bulatov2019integrated}. The normalized GLD always has a value in the range~$[0.0, 1.0]$ and satisfies triangle inequality. It is defined as follows:
\begin{equation}
\label{eq:ngld}
\mathrm{nGLD}(X, Y) = \cfrac{2 \cdot \mathrm{GLD}(X,Y)}{\mathrm{GLD}(X, Y) + \alpha \cdot(|X| + |Y|)}~,
\end{equation}
where $\alpha$ is the maximal component-wise distance between empty and non-empty characters. With a scaled taxicab metric~$\rho_C$ \eqref{eq:scaled_taxicab} the value of $\alpha$ is $1$.
	
If the approximation of the expected distance $\hat{\Delta}_n$ \eqref{eq:modelling_estimation} is calculated directly using the computation scheme \eqref{eq:weighted_coarse_modelling}, the two internal sums in~\eqref{eq:weighted_coarse_modelling} compute the approximation of GLD between $R_n$ and $R(X_1, \ldots, X_n, X_i)$. Thus, each computed GLD can then be normalized under the higher-level sum sign if a normalized version of the GLD is used as a metric function~$\rho$.
	
However, to apply the further optimization \eqref{eq:unweighted_contribution_summation} for an unweighted case, an additional assumption needs to be allowed to convert the computed summation of GLD values to a sum of normalized GLD values.
	
Let us denote as $G_n$ the sum of GLD between the current result $R_n$ and the modelled candidates for the next combined result: $G_n = \sum_{i=1}^n \mathrm{GLD}(R_n, R(X_1, \ldots, X_n, X_i))$. In order to be able to apply the optimization \eqref{eq:unweighted_contribution_summation} we will introduce the following approximation:
	
{\bf Approximation 3:} naive normalization. the GLD constituents of $G_n$ may be normalized after summation. Given approximation 1 (naive alignment) the lengths of $R_n$ and~$R(X_1, \ldots, X_n, X_i)$ are both equal to $S_n$, thus approximation 3 can be expressed as:
\begin{equation}
\label{eq:ngld_local_linearity}
\sum\limits_{i=1}^n \mathrm{nGLD}(R_n, R(X_1, \ldots, X_n, X_i)) \approx \cfrac{2 G_n}{G_n + 2 S_n}.
\end{equation}
	
Since, given the approximation 2 (naive Levenshtein), the value of $\mathrm{GLD}(R_n, R(X_1, \ldots, X_n, X_i))$ is not higher than~$S_n$, the approximation 3 in its essence relies on an assumption of local linearity of function $f(t) = (2t) / (t + 2)$ in the range~$t\in[0.0, 1.0]$. Using the optimization \eqref{eq:unweighted_contribution_summation} and balanced binary search trees we can efficiently compute the approximation of $G_n$, and then use the approximate equation \eqref{eq:ngld_local_linearity} to convert it to a sum of normalized GLD between the current result and the modelled candidates for the next one, required to compute the estimation $\hat{\Delta}_n$.

\section{Experimental evaluation} \label{sec:experiments}
	
In this section, we provide results of an experimental evaluation of the proposed optimization against the direct application of method evaluated in \cite{bulatov2019integrated}. The source code and data necessary to reproduce the experiments are available at the following link: \url{https://github.com/SmartEngines/stoppers_modelling}.
	
\subsection{Experimental setting}
	
The proposed approximations were tested on two different text recognition tasks: the recognition of text fields of identity documents in a video stream, and the recognition of arbitrary text in videos.

For the first task we used open datasets \mbox{MIDV-500} \cite{midv500} and \mbox{MIDV-2019} \cite{midv2019} which contain video sequences of identity documents of various types. The provided ground truth contains coordinates of the document on each frame, for each unique document there are coordinates of each text field and their correct values. The experimental setting closely followed the one described in \cite{bulatov2019integrated} (which only considered \mbox{MIDV-500}) in order to provide a just comparison of the methods. As in \cite{bulatov2019integrated} we considered only frames which fully contain the document boundaries, and to avoid normalization effects each clip was repeated in a loop until the common length of $30$ was reached. Four text field groups were analyzed: document numbers, numerical dates, names written in the Latin alphabet, and machine-readable zone lines. In total there were~$2239$ evaluated clips of MIDV-500 dataset and $992$ clips of \mbox{MIDV-2019} dataset. Each text field image was cropped with the resolution of $300$~DPI and with margins equal to $30\%$ of the smallest text field bounding box side, then recognized using a text string recognition subsystem of Smart IDReader document recognition solution \cite{recstr_access}, thus obtaining string recognition results in form~\eqref{eq:ocrstring}.

	
	

For evaluating the stopping rules on the task of arbitrary text recognition we used a training subset for the Text in Videos task of the ICDAR 2015 Competition on Robust Reading~(IC15-Train)~\cite{7333942} and the YouTube Video Text dataset~(YVT)~\cite{6836024}. From both datasets, we selected text objects with only alphanumeric characters and which were present on at least $30$ frames, then split the video sequences for each object into clips of exactly $30$ frames. Thus, a total of $851$ clips were produced from IC15-Train dataset, and $409$ clips from YVT dataset. Each text object was then cropped using the coordinates provided in the ground truth in the original resolution and recognized using the pre-trained model described in \cite{9010273} with thin-plate spline~(TPS) transformation, ResNet feature extraction,~BiLSTM sequence modelling, and attention-based sequence prediction. The recognition results in form~\eqref{eq:ocrstring} were extracted by removing the softmax outputs corresponding to ``start'' and ``end-of-sentence'' tokens and normalization of the remaining character estimation values.

For both tasks the per-frame text recognition results were combined using the algorithm described in \cite{vestnik_integration} without weights, and a normalized~GLD \cite{ngld_yujian} was used as a metric~$\rho$ between recognition results. The value of $\delta$ parameter in the expected distance estimation~\eqref{eq:modelling_estimation} was taken to be~$0.1$.
	
For each experiment, three methods were evaluated:
\begin{enumerate}
	\item {\em Base method \cite{bulatov2019integrated}.} Estimation value $\hat{\Delta}_n$ is computed using direct modelling \eqref{eq:modelling_estimation}. The complexity of computing the estimation on stage $n$ is $O(nS_nK(S_n+M))$.
	\item {\em Method A.} Estimation is computed using an approximate modelling using direct summation \eqref{eq:weighted_coarse_modelling} with $\Delta_{ijk}$ computed using a simplified expression \eqref{eq:unweighted_contribution} for unweighted case. The complexity of computing the estimation on stage $n$ is $O(nS_nK)$.
	\item {\em Method B.} Estimation is computed using an approximate modelling using a summation scheme optimization~\eqref{eq:unweighted_contribution_summation} with balanced binary search trees. Conversion to normalized GLD is performed using the approximation~3~\eqref{eq:ngld_local_linearity} (naive normalization, see section \ref{sec:ngld}). Treaps with random priorities \cite{Blelloch:1998:FSO:277651.277660} were used as balanced binary search trees due to implementation simplicity. The complexity of computing the approximation of $\hat{\Delta}_n$ on stage~$n$ is~$O(S_nK \log n)$.
\end{enumerate}
	
\subsection{Comparing the distance estimation value}
	
\begin{figure}[t]
	\centering
		\includegraphics[width=\linewidth]{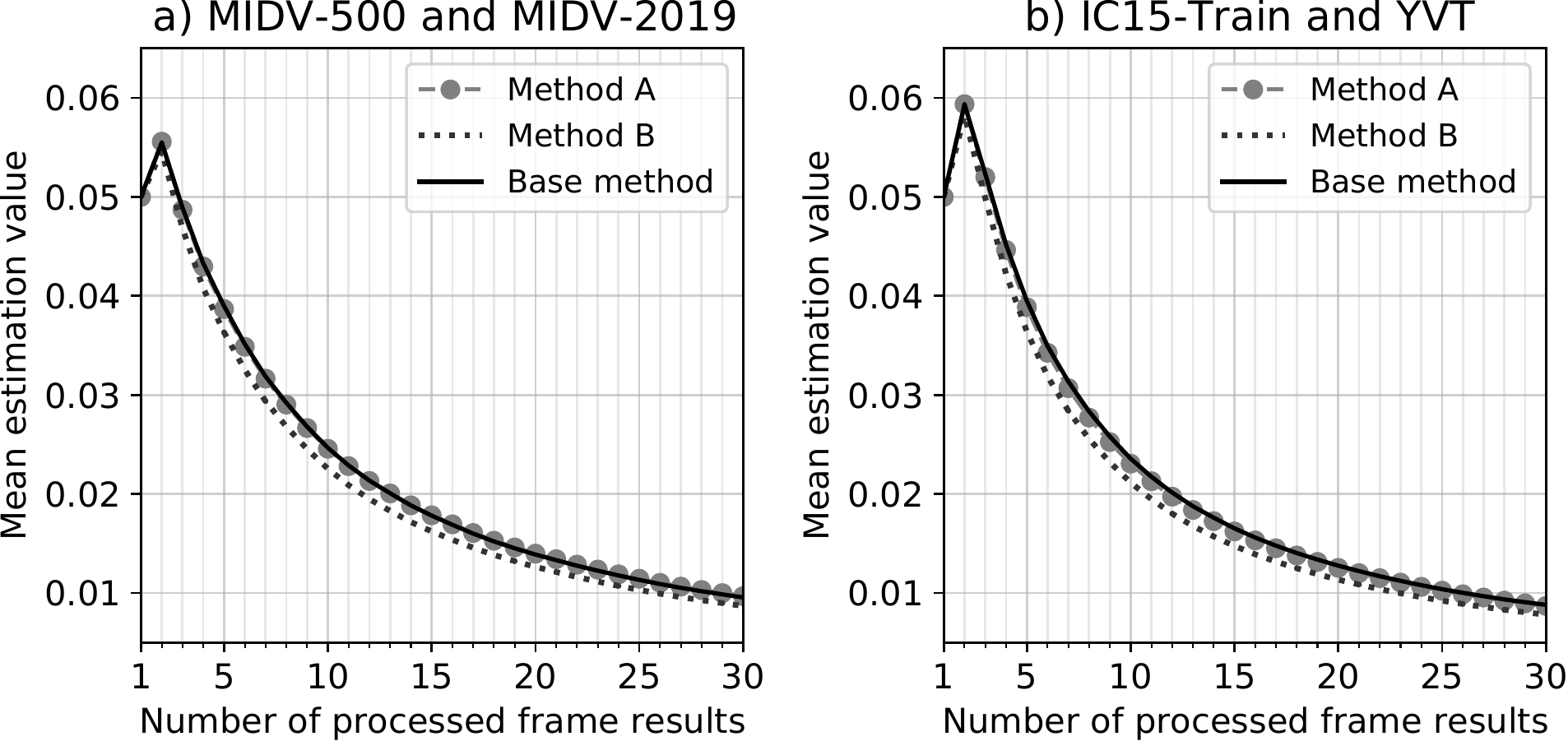}
	\caption{Mean estimation value $\hat{\Delta}_n$ computed using the three evaluated stopping methods on each stage: (a) fields from MIDV-500 and MIDV-2019, (b) text objects from IC15-Train and YVT}
	\label{fig:estimations_comparison}
\end{figure}

For the first experiment we compared the value of the estimated expected distance $\hat{\Delta}_n$ from the current combined result to the potential combination result on the next stage. \figurename~\ref{fig:estimations_comparison} illustrates the plot of the mean estimation value on each simulation stage on both datasets for both tasks.

If can be seen from \figurename~\ref{fig:estimations_comparison} that the estimation value calculated with Method A equals on average to the estimation calculated using the full modelling \eqref{eq:modelling_estimation}. This signifies that the approximations 1 and 2 (naive alignment and naive Levenshtein) introduced in subsection \ref{sec:approximations} are justified for efficient approximate computation of the expected distance estimation. The approximation error of~Method B is an effect of the naive normalization \eqref{eq:ngld_local_linearity}.
	
\subsection{Comparing the modelling time}
	
\begin{figure}[t]
    \centering
		\includegraphics[width=\linewidth]{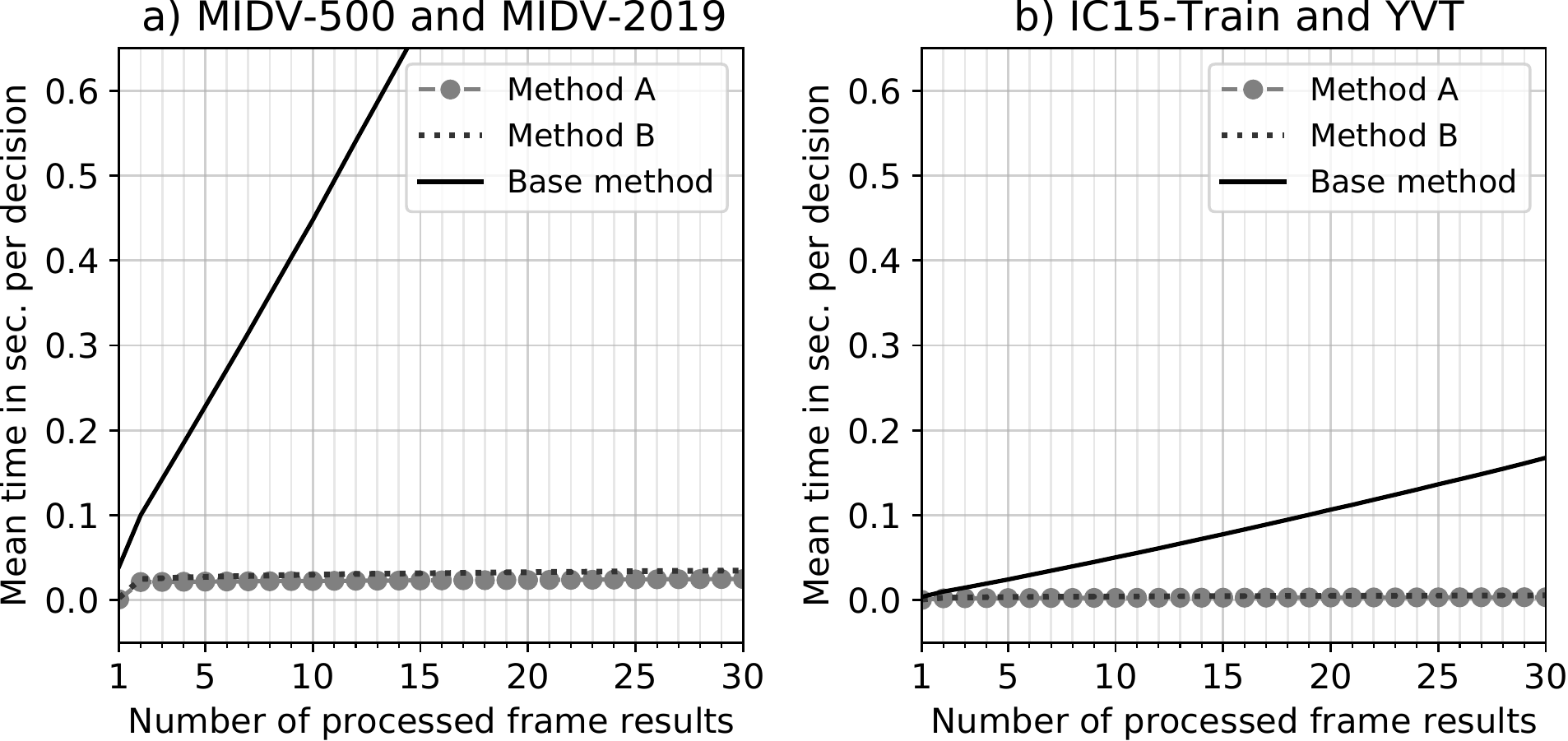}
	\caption{Comparison of the time required to construct the combined result $R_n$ and calculate distance estimation $\hat{\Delta}_n$, for the evaluated methods: (a) fields from MIDV-500 and MIDV-2019, (b) text objects from IC15-Train and YVT}
	\label{fig:timing_comparison}
\end{figure}

To evaluate the computational performance of the proposed approximations we compared the combined time required to produce the result $R_n$ (as the proposed method involve on-the-fly modifications of some internal structures during the results combination process) and calculate on stage~$n$ the estimation of the expected distance from the current combined result $R_n$ to the next result $R_{n+1}$. The combined time to perform such operations for the fields in MIDV-500 and MIDV-2019 are presented in Table~\ref{tbl:timing_midv}, and the same for the text objects in~IC15-Train and YVT is presented in Table~\ref{tbl:timing_titw}. \figurename~\ref{fig:timing_comparison} shows the timing comparison for each stage $n$ for both tasks. Measurements were performed for a single-thread Python prototype implementation~(Python 3.7.4 under Jupyter~6.0.1, AMD~Ryzen~9~3950X, 64Gb RAM, GTX 1050 GPU).
	
\begin{table}[!t]
\renewcommand{\arraystretch}{1.3}
\caption{Time Required to Construct the Combined Result and Calculate the Expected Distance Estimation: Fields~From~MIDV-500~and~MIDV-2019}
\label{tbl:timing_midv}
	\begin{center}
		\begin{tabular}{|p{1.3cm}|p{0.9cm}|p{0.9cm}|p{0.9cm}|p{0.9cm}|p{0.9cm}|}
			\hline\rule{0pt}{2.6ex}
			\multirow{2}{*}{Method} & \multicolumn{5}{c|}{Time on stage $n$ (in seconds)}             \\ \cline{2-6}\rule{0pt}{2.6ex} 
			& $n=5$ & \mbox{$n=10$}    & \mbox{$n=15$}    & \mbox{$n=20$}    & \mbox{$n=25$}    \\ \hline\hline\rule{0pt}{2.5ex} 
			Base \cite{bulatov2019integrated} & $0.228$ & $0.448$ & $0.678$ & $0.906$ & $1.143$ \\ \hline
			\mbox{Method A}                   & $\mathbf{0.022}$ & $\mathbf{0.022}$ & $\mathbf{0.023}$ & $\mathbf{0.024}$ & $\mathbf{0.024}$ \\ \hline
			\mbox{Method B}                   & $0.027$ & $0.030$ & $0.032$ & $0.033$ & $0.034$ \\ \hline
		\end{tabular}
	\end{center}
\end{table}

\begin{table}[!t]
\renewcommand{\arraystretch}{1.3}
\caption{Time Required to Construct the Combined Result and Calculate the Expected Distance Estimation: Text~Objects~From~IC15-Train~and~YVT}
\label{tbl:timing_titw}
	\begin{center}
		\begin{tabular}{|p{1.3cm}|p{0.9cm}|p{0.9cm}|p{0.9cm}|p{0.9cm}|p{0.9cm}|}
			\hline\rule{0pt}{2.6ex}
			\multirow{2}{*}{Method} & \multicolumn{5}{c|}{Time on stage $n$ (in seconds)}             \\ \cline{2-6}\rule{0pt}{2.6ex} 
			& $n=5$ & \mbox{$n=10$}    & \mbox{$n=15$}    & \mbox{$n=20$}    & \mbox{$n=25$}    \\ \hline\hline\rule{0pt}{2.5ex} 
			Base \cite{bulatov2019integrated} & $0.024$ & $0.050$ & $0.078$ & $0.107$ & $0.136$ \\ \hline
			\mbox{Method A}                   & $\mathbf{0.002}$ & $\mathbf{0.003}$ & $\mathbf{0.003}$ & $\mathbf{0.003}$ & $\mathbf{0.003}$ \\ \hline
			\mbox{Method B}                   & $0.004$ & $0.004$ & $0.005$ & $0.005$ & $0.005$ \\ \hline
		\end{tabular}
	\end{center}
\end{table}

It is evident from the data in \figurename~\ref{fig:timing_comparison} that approximate computation of the estimation $\hat{\Delta}_n$ allows to make the stopping decision much quicker in comparison with direct modelling, and that the increase of required computations from stage to stage becomes negligible. As mentioned in section~\ref{sec:unweighted_case}, the difference between the direct summation in~Method A and the optimized summation scheme \eqref{eq:unweighted_contribution_summation} with balanced search trees in Method~B is insignificant, and the latter is even slightly less computationally efficient, due to the data structure overhead.

\subsection{Comparing the stopping method performance}
	
\begin{figure*}[t]
    \centering
		\includegraphics[width=0.775\linewidth]{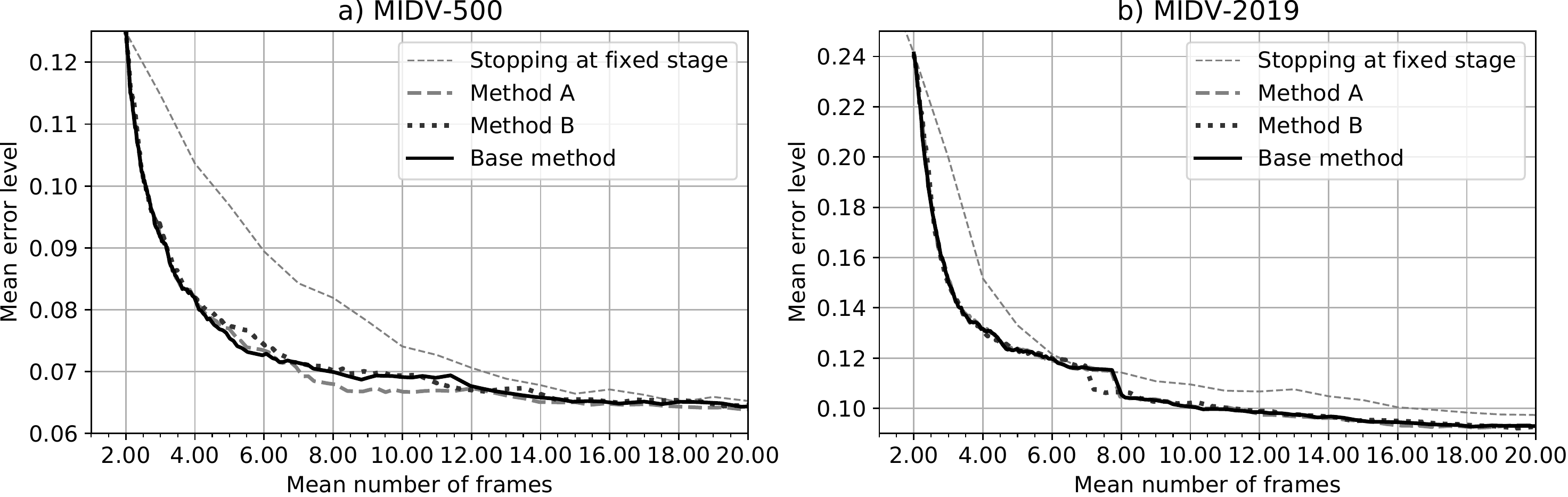}
	\caption{Expected performance profiles of the three evaluated methods: mean distance from the obtained combination result to the correct value against the mean number of processed frames before stopping, with a varied stopping threshold: (a) fields from MIDV-500 dataset, (b) fields from MIDV-2019 dataset}
	\label{fig:epps_comparison_midv}
\end{figure*}
	
\begin{figure*}[t]
    \centering
		\includegraphics[width=0.775\linewidth]{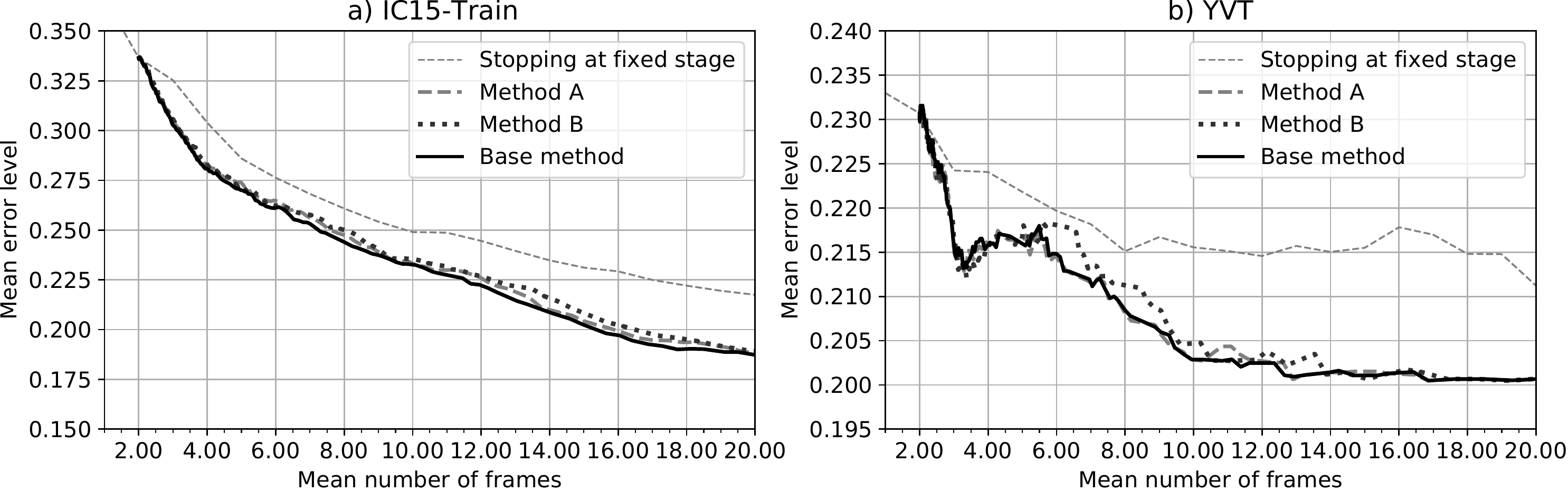}
	\caption{Expected performance profiles of the three evaluated methods: mean distance from the obtained combination result to the correct value against the mean number of processed frames before stopping, with a varied stopping threshold: (a) text objects from IC15-Train, (b) text objects from YVT dataset}
	\label{fig:epps_comparison_titw}
\end{figure*}

The final experiment was aimed at comparing the performance of the resulting modified stopping methods. A ``good'' stopping method should be able to achieve lower error level given a fixed mean number of frames, or, respectively, the lower mean number of processed frames given a fixed mean error level. A convenient method of comparing stopping methods is to compare their expected performance profiles, which are used for anytime algorithms analysis \cite{Zilberstein_1996}. Such performance profile can be obtained by plotting the mean error level (in terms of distance to the correct result) of the combined result at the stopping stage and the mean number of the stopping stage, while varying observation cost value. The lower position of the plotted curve indicates the greater performance of the stopping method. The observation cost in the analyzed methods corresponds to the threshold with which the estimation~$\hat{\Delta}_n$ is compared when making the stopping decision. 
	
\figurename~\ref{fig:epps_comparison_midv} illustrates the expected performance profiles of the three evaluated methods, as well as a baseline stopping method which stops the process after a fixed stage, for text fields in the MIDV-500 and MIDV-2019 datasets. The varied parameter for the baseline stopping method is the number of the stage on which it should stop. The corresponding profiles for the arbitrary text recognition task are presented in \figurename~\ref{fig:epps_comparison_titw}.
	
It can be seen from \figurename~\ref{fig:epps_comparison_midv} and \ref{fig:epps_comparison_titw} that the approximations introduced in sections \ref{sec:approximations} and \ref{sec:ngld} had almost no effect on the performance of the stopping method in terms of the achieved error level and required number of observations. There is a slight disadvantage of the~Method~B against the Method A, which could be an effect of utilizing the naive normalization approximation~3~\eqref{eq:ngld_local_linearity} when using a normalized GLD as a metric function~$\rho$. Given the significant advantage in computational performance, it can be concluded that the proposed approximate modelling method is more favourable for stopping the text recognition process in a video stream, applicable for both document analysis, and for recognition of an arbitrary text.

\section{Conclusion}
	
In this paper, we described the problem of making the stopping decision efficiently for text string recognition in a video stream. An overview was given for an existing method based on modelling of the next combined result. The disadvantage of this method is its high computational complexity of the modelling, resulting in a slow decision making process. In the paper, we proposed two approximate modelling schemes, one which is applicable for a general weighted case, and one for unweighted case. Both optimizations were evaluated on two open datasets and in the same experimental setting as the previously introduced method. 

It can be seen from the presented experimental evaluation that the assumptions and approximations introduced in the paper had almost no effect on the performance of the stopping method in terms of the achieved mean error level and a mean number of consumed observations. At the same time, the proposed computational schemes have significantly higher computational performance. The obtained results were consistent for two different text recognition tasks: identity documents recognition, and arbitrary text recognition, each with a different text recognition algorithm.

Optimal stopping of object recognition in a video stream is an important pattern recognition problem. The role of stopping methods as components of modern video recognition systems is very valuable, not only from the perspective of the real-time interaction with the user but from the more general perspective of being able to achieve more accurate recognition results with adequate response time.

\section*{Acknowledgment}

This work is partially financially supported by the Russian Foundation for Basic Research (projects 19-29-09055 and \mbox{18-07-01387}).



\bibliographystyle{IEEEtran}
\bibliography{IEEEabrv,bibliography}
%

\end{document}